\documentclass[10pt,twocolumn,letterpaper]{article}

\usepackage{cvpr}              

\usepackage[accsupp]{axessibility}
\usepackage{marvosym}
\usepackage{graphicx}
\usepackage{amsmath}
\usepackage{amssymb}
\usepackage{booktabs}
\usepackage{colortbl}
\usepackage{multirow}
\usepackage{xcolor}
\definecolor{cvprblue}{rgb}{0.21,0.49,0.74}
\definecolor{iblue}{rgb}{0.06, 0.75, 1.0}

\newcommand\blfootnote[1]{%
  \begingroup
  \renewcommand\thefootnote{}\footnote{#1}%
  \addtocounter{footnote}{-1}%
  \endgroup
}
\definecolor{cvprblue}{rgb}{0.21,0.49,0.74}
\usepackage[pagebackref,breaklinks,colorlinks,allcolors=cvprblue]{hyperref}

\def\paperID{30889} 
\def\confName{CVPR}
\def\confYear{2026}

\title{Captain Safari: A World Engine with Pose-Aligned 3D Memory}

\author{
Yu-Cheng Chou\textsuperscript{1} \quad
Xingrui Wang\textsuperscript{1} \quad
Yitong Li\textsuperscript{2} \quad
Jiahao Wang\textsuperscript{1} \quad
Hanting Liu\textsuperscript{1} \quad \\
Cihang Xie\textsuperscript{3} \quad
Alan Yuille\textsuperscript{1} \quad
Junfei Xiao\textsuperscript{1\Letter}\\
\textsuperscript{1}Johns Hopkins University \quad
\textsuperscript{2}Tsinghua University \quad
\textsuperscript{3}UC Santa Cruz
}

\begin{document}
\setlength{\abovedisplayskip}{6pt}
\setlength{\belowdisplayskip}{6pt}
\setlength{\abovedisplayshortskip}{6pt}
\setlength{\belowdisplayshortskip}{6pt}

\twocolumn[{%
\renewcommand\twocolumn[1][]{#1}%
\maketitle
\vspace{-2.5em}
\centering
\url{https://johnson111788.github.io/open-safari/}
\begin{center}
\includegraphics[width=0.95\linewidth]{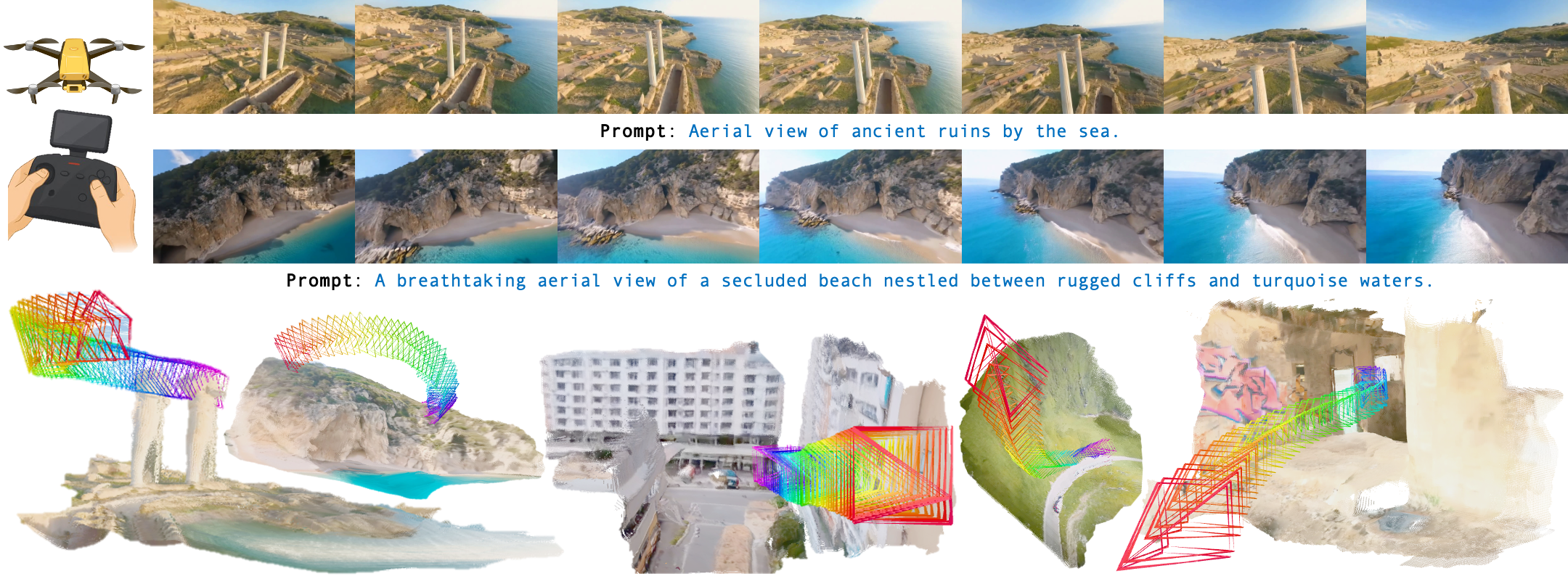}
\captionof{figure}{
\textbf{Captain Safari} is a pose-aware world engine that generates long-horizon, 3D-consistent FPV videos from any user-specified camera trajectory.
By retrieving pose-aligned world memory, it keeps geometry stable across large viewpoint changes and reconstructs crisp, well-formed structures while faithfully tracking aggressive 6-DoF motion.
}
\label{fig:teaser}
\end{center}
}]

\blfootnote{\textsuperscript{\Letter} Corresponding author: Junfei Xiao (xiaojf97@gmail.com)}

\begin{abstract}
World engines aim to synthesize long, 3D-consistent videos that support interactive exploration of a scene under user-controlled camera motion. 
However, existing systems struggle under aggressive 6-DoF trajectories and complex outdoor layouts: they lose long-range geometric coherence, deviate from the target path, or collapse into overly conservative motion. 
To this end, we introduce \emph{Captain Safari}, a pose-conditioned world engine that generates videos by retrieving from a persistent world memory.
Given a camera path, our method maintains a dynamic local memory and uses a retriever to fetch pose-aligned world tokens, which then condition video generation along the trajectory. 
This design enables the model to maintain stable 3D structure while accurately executing challenging camera maneuvers.

To evaluate this setting, we curate \emph{OpenSafari}, a new in-the-wild FPV dataset containing high-dynamic drone videos with verified camera trajectories, constructed through a multi-stage geometric and kinematic validation pipeline. 
Across video quality, 3D consistency, and trajectory following, \emph{Captain Safari} substantially outperforms state-of-the-art camera-controlled generators. 
It reduces MEt3R from 0.3703 to 0.3690, improves AUC@30 from 0.181 to 0.200, and yields substantially lower FVD than all camera-controlled baselines.
%
More importantly, in a 50-participant, 5-way human study where annotators select the best result among five anonymized models, \textbf{67.6\%} of preferences favor our method across all axes.
Our results demonstrate that pose-conditioned world memory is a powerful mechanism for long-horizon, controllable video generation and provide \emph{OpenSafari} as a challenging new benchmark for future world-engine research.
\end{abstract}    
\section{Introduction}
\label{sec:intro}
Simulating coherent 3D worlds through controllable video generation has long been a foundational challenge for augmented reality, embodied AI, and virtual agents~\citep{wu2025geometry,huang2025memory,gao2025adaworld,wu2025vmoba,wiles2020synsin,motionprompt,cami2v,li2025realcam,PoseTraj,CameraCtrl,lu2024genex,wang2025evoworld,liang2025wonderland,yu2025wonderworld,ni2025wonderturbo,hu2021worldsheet,yu2021pixelnerf,voyager,Gen3c}.
Classical game engines and physics simulators offer explicit geometry and precise control, but require heavy manual authoring and expensive computation~\citep{ramakrishnan2021habitat,chang2017matterport3d,savva2019habitat}.
Moreover, while they may achieve visual realism in specialized domains, they still fall short in capturing the richness and diversity characteristic of real world, such as natural scenes~\citep{pun2023neural,tan2021scenegen,zhou2024scenex}.
In contrast, modern video diffusion models synthesize high-fidelity, diverse videos from text or images, yet typically operate as feed-forward clip generators without persistent world state: 
\textit{they struggle with long-range 3D consistency, complex trajectory following, and faithful reconstruction of diverse scenes}~\cite{melnik2024video,kuang2024collaborative}.
In this work, we move toward bridging this gap with \emph{Captain Safari}, a world engine that enables pose-conditioned modeling of 3D-consistent and diverse environments, surpassing the limitations of traditional game engines in terms of generality, diversity, and interactivity.

Contemporary video world models face three intertwined challenges. 
First, \emph{long-horizon consistency} is limited by the temporal window of context frames; models often “forget” distant scenery or violate spatial coherence, leading to abrupt appearance changes that break the realism and continuity of the generated environment~\citep{wu2025vmoba,gao2025adaworld,huang2025memory}. 
Second, achieving \emph{complex camera maneuvers under strict 3D consistency} remains difficult: existing pose- or trajectory-conditioned methods typically work well only for slow, near-forward motions~\cite{PoseTraj,yang2025omnicam,shuai2025free}. 
When the path involves fast 6-DoF movement, strong parallax, or sharp turns, models exhibit a trade-off—either dampening motion and restricting viewpoint changes to preserve geometry, or committing to the requested path at the cost of distortions, flicker, and structural drift. 
Third, current approaches underrepresent \emph{complex outdoor layouts}. Much of the works focuses on structured, constrained settings (e.g., indoor tours, driving scenes, or real-estate videos), and models are seldom stress-tested in in-the-wild FPV scenarios where the camera weaves around buildings, vegetation, and varied terrain with substantial parallax~\cite{zhou2018stereo,lu2024wovogen,zhou2024simgen,cao2021visdrone}. 
As a result, methods that look competitive in simplified environments often fail to preserve geometry and appearance when confronted with truly diverse, complex outdoor scenes.

To address these issues, we introduce \emph{Captain Safari}, a pose-aware world engine that explicitly maintains a persistent notion of world state to uphold \emph{long-horizon} 3D consistency across strong parallax.
Because storing and propagating a full long-term state is computationally prohibitive, we develop a retrieval mechanism that \emph{selects and aggregates} only the most informative scene cues, thereby providing strong geometric guidance without incurring prohibitive cost.
Crucially, this retrieval is \emph{pose-aware}: given the target camera pose, it assembles a pose-aligned world prior that steers the generation process, enabling accurate tracking of \emph{aggressive camera maneuvers} while preserving 3D-consistent structure in complex environments.

Furthermore, to close the gap in \emph{complex outdoor layouts} and \emph{aggressive camera motion}, we curate \emph{OpenSafari}, a large-scale dataset of high-dynamic FPV drone videos with verified camera poses.
Much of the literature targets structured, constrained settings (e.g., indoor tours, driving or real-estate videos), and even outdoor datasets typically feature slow, near-forward motion.
In contrast, \emph{OpenSafari} comprises in-the-wild FPV flights that weave around buildings and vegetation across uneven terrain, exhibiting large parallax, rapid 6-DoF maneuvers, and sharp viewpoint changes.
Paired with verified camera trajectories, these videos present diverse, cluttered outdoor scenes and long-range motion, challenging models to maintain 3D consistency while faithfully tracking complex maneuvers.

We evaluate \emph{Captain Safari} along three axes: \emph{video quality}, \emph{3D consistency}, and \emph{trajectory following}.
Across these criteria, our method consistently outperforms contemporary camera-controlled video generators on \emph{OpenSafari}: 
Table~\ref{tab:main_results} reports clear gains in 3D consistency and accurate tracking under complex maneuvers, while maintaining strong perceptual quality.
Importantly, a large-scale human study (Table~\ref{tab:human}) shows that \emph{Captain Safari} receives \textbf{67\%} of votes in five-way comparisons, indicating that the improvements are perceptually salient.
Qualitative comparisons (Fig.~\ref{fig:qual} and Fig.~\ref{fig:reconstruction}) further demonstrate stable geometry under long-range path and faithful adherence to sharp 6-DoF camera turns in cluttered outdoor scenes.


In summary, our contributions are:
\begin{enumerate}
    \item We present \emph{Captain Safari}, the first camera-controlled video generation method to enforce long-horizon 3D consistency while tracking aggressive FPV maneuvers.
    \item We propose a \emph{pose-guided, long-horizon retrieval} that efficiently reconciles strict 3D consistency with accurate tracking of complex maneuvers.
    \item We curate \emph{OpenSafari}, a large-scale in-the-wild FPV dataset with verified camera poses, featuring diverse, cluttered outdoor scenes and rapid 6-DoF motion that stress-test geometry-consistent camera control.
    \item In \emph{OpenSafari}, our pose-aware retrieval notably improves video quality, 3D consistency, and trajectory alignment, also achieving a \textbf{67\%} human preference rate.

\end{enumerate}


\begin{figure*}[t]
\centering
\includegraphics[width=\linewidth]{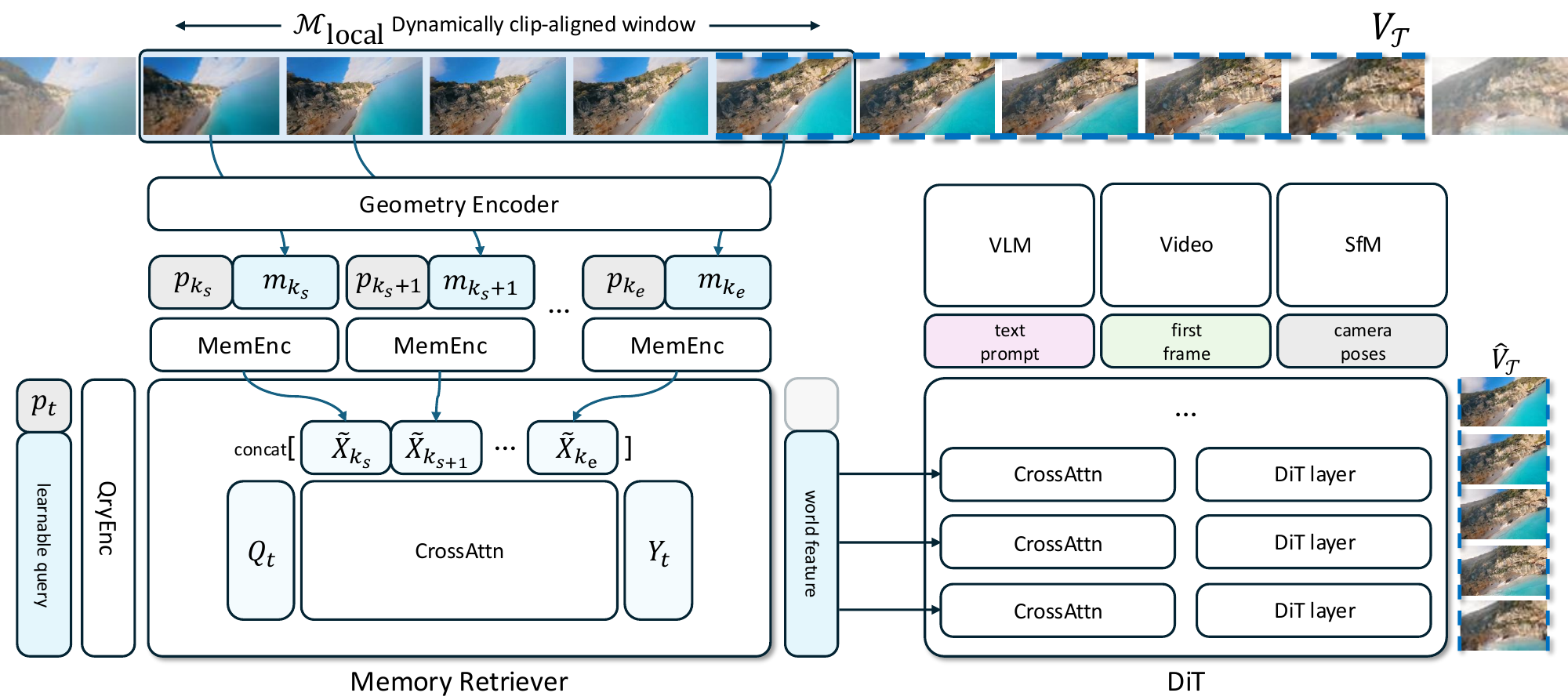}
\vspace{-1.5em}
\caption{
\textbf{Method overview.} \emph{Captain Safari} builds a local world memory and, given a query camera pose, retrieves pose-aligned tokens that summarize the scene. 
These tokens then condition video generation along the user-specified trajectory, preserving a stable 3D layout.\vspace{-1em}
}
\label{fig:pipeline}
\end{figure*}

\section{Related Work}

\subsection{3D-Consistent World Models}

Early image-to-3D approaches reconstruct geometry indirectly via multi-view consistency or implicit fields, but often fail to maintain coherent structure across large view changes~\citep{wiles2020synsin,yu2021pixelnerf,liu2023zero,hu2021worldsheet}. 
Recent efforts integrate 3D reasoning into the generative process. DiffusionGS~\citep{cai2024baking} injects Gaussian Splatting into the diffusion denoiser, enforcing view consistency and enabling single-stage, scalable 3D generation.
GenEx~\citep{lu2024genex} and EvoWorld~\citep{wang2025evoworld} extends this idea from static reconstruction to dynamic world creation, generating explorable 360° panoramic environments grounded in physical priors.
Complementary to these generative reconstructions, Geometry Forcing~\citep{wu2025geometry} and Memory Forcing~\citep{huang2025memory} explicitly couple training signals with geometric supervision and spatio-temporal memory, ensuring consistency during long rollouts.  
Meanwhile, open-world models such as Wonderland~\citep{liang2025wonderland}, WonderWorld~\cite{yu2025wonderworld}, WonderTurbo~\citep{ni2025wonderturbo}, 
RELIC~\cite{hong2025relic}, and
WorldPlay~\cite{sun2025worldplay} further integrate geometry-indexed or adaptive memories to maintain persistent world states across interactions.  
However, these approaches still use implicit, clip-bound memories, or explicitly maintain heavy point clouds~\cite{wu2025video}, whereas our decoupled retriever keeps the condition compact at fixed cost.


\subsection{Camera Controlled Video Generation}

Early T2V/I2V models learned camera motion implicitly and struggle to reliably repeat explicit trajectories~\citep{hou2024training,wang2024motionctrl,zhou2025latent}.
Recent work such as CameraCtrl~\citep{CameraCtrl} treats camera parameters as explicit conditions, encoding camera extrinsics and trajectories or enforcing path constraints—to improve controllability and accuracy~\citep{mou2024revideo,videocomposer,tc4d,tora,CamCo}. 
MotionPrompting~\citep{motionprompt} implement compositional control by point-track conditioning and MotionPro~\citep{MotionPro} use path-alignment losses that lower rotational and translational error; training-free control is also achieved by fitting a lightweight pointcloud and using a noise-layout prior to steer denoising~\citep{hou2024training}.
Scene-preserving geometric priors further strengthen clip-level consistency. Cami2V\citep{cami2v} treats camera pose as a physical prior and exploits epipolar and multiview constraints; RealCam-I2V~\citep{li2025realcam} recovers metric depth with DepthAnything v2~\citep{yang2024depth} to reconstruct a scale-stable scene; PoseTraj\citep{PoseTraj} employs pose-aware pretraining to obtain rotation-aligned motion. Compared with parameter-only conditioning, these priors reduce within-clip layout drift and better preserve local geometry under view changes.
Further, recent work links camera control with world modeling. CVD~\citep{cvd}, Cavia~\citep{xu2024cavia}, and WoVoGen~\citep{lu2024wovogen} jointly synthesize multi-view and multi-trajectory videos from a shared scene representation, enforcing cross-path consistency. Meanwhile, methods that conditioning from explicit renderable 3D representations (e.g., 3D Gaussians) can anchor geometry, improve cross-view 3D consistency and path adherence ~\citep{viewcrafter,Gen3c,trajectorycrafter,worldexplorer,voyager}. 
However, these approaches typically build one-off 3D scenes, whereas we unify long-horizon camera control with a persistent pose-indexed world memory shared across trajectories.


\section{Captain Safari}
We introduce \emph{Captain Safari}, a memory-guided video generation framework. Sec.~\ref{sec:memory} presents an implicit world memory for stable scene representation, while Sec.~\ref{sec:retrieval} describes a pose-conditioned retrieval system that maps camera views to world tokens, guiding a DiT-based generator for coherent outputs along arbitrary trajectories.

\subsection{Implicit Memory of World Geometry}\label{sec:memory}
\noindent\textbf{Problem setup.}
We represent a video as
\(V = \{I_t\}_{t=0}^{T}\), where \(I_t\) is the frame at time step \(t\).
On the same time axis we define camera poses
\(\mathcal{C} = \{(R_t, T_t)\}_{t=0}^{T}\)
and obtain a 3D-aware memory feature
\(m_{t}\)
at each time step \(t\) using a pretrained geometry encoder.
All memory features form a global memory bank
\(\mathcal{M} = \{m_{t}\}_{t=0}^{T}\).

Given a text prompt \(p\), the camera poses \(\mathcal{C}\), and a target clip
time step \(\mathcal{T} = [t_0,t_1]\),
together with its associated local world memory
\(\mathcal{M}_{\text{local}} \subset \mathcal{M}\),
our goal is to synthesize a video segment \(\hat V_{\mathcal{T}}\) that
(i) aligns with \(p\),
(ii) respects the prescribed poses \(\{(R_t, T_t)\}_{t \in \mathcal{T}}\),
and (iii) maintains a coherent 3D world across viewpoints.

\noindent\textbf{Local world memory.}
Directly conditioning on the full memory bank \(\mathcal{M}\) for every clip would be computationally expensive and dominated by temporally distant observations.
Instead, for each target clip time step \(\mathcal{T} = [t_0,t_1]\) we define a \emph{local} memory
\(\mathcal{M}_{\text{local}} = \{ m_{\tau} \mid \tau \in [k_s, k_e] \}\)
whose endpoints are sampled under
\begin{equation}
\begin{aligned}
t_0 - L \;\le\; &k_s \;\le\; t_0,\\
\max(k_s,\, t_0){+}1 \;\le\;& k_e \;\le\; \min(k_s{+}L,\, t_1),
\end{aligned}
\label{eq:local-memory-constraints}
\end{equation}
where \(L\) is a fixed bound and all time steps are integers.
These constraints enforce that:
(i) the memory window starts at most \(L\) seconds before the clip entrance \(t_0\), tying it to nearby observations;
(ii) its duration is at most \(L\), which keeps the conditioning set compact; and
(iii) its end time \(k_e\) always touches or overlaps \(t_0\) while remaining within \([t_0,t_1]\), ensuring that each clip is supported by a temporally compatible world prior.
All \(\mathcal{M}_{\text{local}}\) are constructed as such dynamic clip-aligned window of the shared bank \(\mathcal{M}\), so neighboring clips naturally share overlapping memory entries, constraining computation while coupling their generations to a 3D-consistent underlying world. 

\noindent\textbf{Pose-retrieved memory.}
Within a given clip time step \(\mathcal{T}\), we treat the local memory \(\mathcal{M}_{\text{local}}\) as a static hypothesis of the surrounding world built from key frames.
Each time step \(\tau\) provides a pose token \(p_\tau\) (derived from \((R_\tau, T_\tau)\)) and a set of 3D-aware memory tokens \(m_{\tau,1},\dots,m_{\tau,M}\).
The collection \(\{(p_\tau, m_{\tau,1:M})\}_\tau\) forms an implicit world table: pose token indicates \emph{where} the camera has observed the scene, while memory tokens encode \emph{what} the world looks like from those configurations.
For any target time step \(t \in \mathcal{T}\), we derive its camera pose to a query pose token \(p_t\), embed it as \(q_t = \phi_p(p_t)\), and use a dedicated retrieval module to read from this static table in a pose-dependent manner.
Concretely, \(q_t\) is concatenated with a bank of learnable query tokens and processed into retrieval queries, which perform cross-attention over the encoded memory \(\tilde X^{\text{mem}}\) (defined in Sec. \ref{sec:retrieval}), yielding a set of world tokens
\begin{equation}
w_t = \mathrm{Agg}\Big(\mathrm{CrossAttn}(Q_t,\; \tilde X^{\text{mem}})\Big)
\end{equation}
corresponding to the updated learnable queries.
These pose-aligned world tokens \(w_t\) are directly used as the reconstructed memory at pose \(t\).
Thus, all frames in \(\mathcal{T}\) access local memory through pose-conditioned queries instead of raw time indices, encouraging multi-view observations to remain tied to a consistent static 3D world.

\begin{figure*}[t]
\centering
\includegraphics[width=0.9\linewidth]{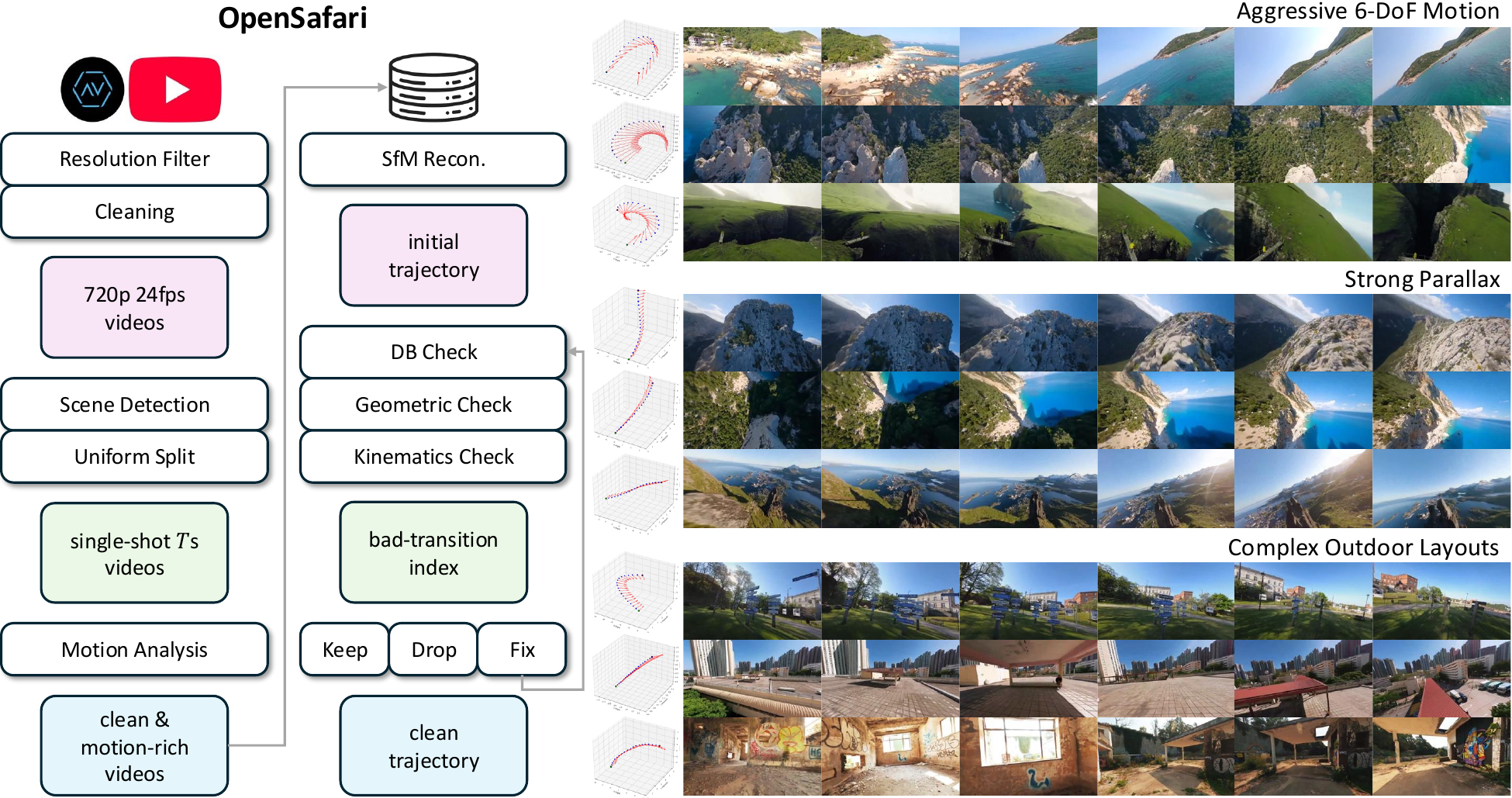}
\vspace{-0.3em}
\caption{\textbf{\emph{OpenSafari}}. A new in-the-wild FPV dataset with rigorously verified camera trajectories, designed to stress-test geometry-consistent, camera-controllable video generation. 
We curate clips through a compact, multi-stage pipeline that filters, reconstructs, and verifies trajectories, yielding clean, motion-rich videos with reliable camera paths.\vspace{-1em}}
\label{fig:data_pipeline}
\end{figure*}

\subsection{Memory Retrieval and Conditioning}\label{sec:retrieval}
\noindent\textbf{Memory retriever design.}
As shown in Figure~\ref{fig:pipeline}, given the local memory, we represent each time step \(\tau\) by a pose token \(p_\tau\) and its associated memory tokens \(m_{\tau,1:M}\).
Our retriever is designed to (i) jointly encode pose--memory pairs into a coherent world representation, and (ii) extract, for any query pose, a compact set of pose-aligned tokens that summarize the most relevant parts of this local world.

We first embed pose and memory features into a shared space and form a joint sequence per time step
\begin{equation}
\hat{X_\tau} = \big[\phi_p(p_\tau),\; \phi_m(m_{\tau,1}),\dots,\phi_m(m_{\tau,M})\big],
\end{equation}
where \(\phi_p\) and \(\phi_m\) denote learnable embeddings for pose and memory tokens, respectively.
A stack of transformer blocks (\(\mathrm{MemEnc}\)) with 3D RoPE~\cite{su2024roformer} refines these sequences
\begin{equation}
\tilde X_\tau = \mathrm{MemEnc}\!\big(\hat{X_\tau}\big),
\end{equation}
and we obtain encoded memory \(\tilde X^{\text{mem}}\) by concatenation
\begin{equation}
\tilde X^{\text{mem}} = \big[\tilde X_{k_s},\dots,\tilde X_{k_e}\big],
\end{equation}
optionally masked to exclude padded or non-key entries.

For a target time step \(t\), we derive the query pose token \(p_t\), embed it as \(q_t = \phi_p(p_t)\), and concatenate it with \(M\) learnable query tokens \(r_1,\dots,r_M\),
\begin{equation}
\hat{Q_t} = [q_t, r_1,\dots,r_M].
\end{equation}
This sequence is refined by transformer blocks sharing the same architecture as \(\mathrm{MemEnc}\), denoted as \(\mathrm{QryEnc}\), yielding pose-aware retrieval queries
\begin{equation}
Q_t = \mathrm{QryEnc}\!\big(\hat{Q_t}\big).
\end{equation}
We then perform cross-attention from \(Q_t\) to \(\tilde X^{\text{mem}}\),
\begin{equation}
Y_t = Q_t \;+\; \mathrm{CrossAttn}(Q_t,\; \tilde X^{\text{mem}}),
\end{equation}
and take the subset of tokens in \(Y_t\) corresponding to the learnable queries as the retrieved world tokens
\begin{equation}
w_t = [w_{t,1},\dots,w_{t,M}],
\end{equation}
which form a pose-aligned world tokens for time \(t\).
During training, a linear head maps \(w_t\) back to the original memory space to reconstruct the target memory tokens at the query pose.
Stacking multiple retrieval blocks iteratively refines both the queries and the retrieved tokens, enabling the model to softly route each query pose to the most relevant subset of past observations, instead of relying on a rigid temporal neighborhood or a single nearest frame.

\noindent\textbf{Memory-conditioned DiT.}
For a given target clip time step \(\mathcal{T}\), the retriever consumes \(\mathcal{M}_{\text{local}}\) and the query pose \(p_t\) and outputs a pose-aligned set of world tokens
\(
w_{t} \in \mathbb{R}^{M \times d_m},
\)
which summarize the static local world relevant to this segment.
These tokens are mapped into the DiT hidden space by the memory embedding MLP
\begin{equation}
W_{\mathcal{T}} = \phi_{w}(w_{t}) \in \mathbb{R}^{M \times D}.
\end{equation}

The latent clip is encoded as a single spatio-temporal token sequence
\(
Z \in \mathbb{R}^{L_z \times D},
\)
obtained by patchifying all frames in \(V_{\mathcal{T}}\).
At each DiT layer \(l\), we first apply self-attention over the full sequence and then inject the world tokens through a dedicated memory cross-attention:
\begin{equation}
Z^{(l+1)} \;=\; Z^{l} \;+\; \mathrm{CrossAttn}(
    Z^{l},\;
    W_{\mathcal{T}},\;
    W_{\mathcal{T}}
).
\end{equation}
The clip-level world tokens \(W_{\mathcal{T}}\) are reused as keys and values across all layers, providing a stable, 3D-consistent prior that shapes the denoising of every spatio-temporal token.
\section{OpenSafari}

\subsection{Video Data Curation}
Existing camera-conditioned datasets do not match our target regime.
RealEstate10K~\cite{zhou2018stereo} focuses on slow, mostly indoor real-estate walkthroughs with gentle motion and clean, quasi-static scenes, while Minecraft~\cite{baker2022video} is a synthetic voxel world with simplified geometry and engine-constrained dynamics.
Neither captures aggressive, in-the-wild 6-DoF drone flight with strong parallax, large elevation changes, and complex outdoor layouts that truly stress long-horizon 3D consistency.
We therefore propose \emph{OpenSafari}, a new dataset of real-world FPV-style drone videos with verified camera trajectories tailored to this challenging setting.

We construct Safari-FPV from FPV-style drone videos collected on AirVuz\footnote{\url{https://www.airvuz.com/}} and YouTube\footnote{\url{https://www.youtube.com/}}, and retain only clips that pass a strict multi-stage preprocessing pipeline.
As shown in Figure~\ref{fig:data_pipeline}, we:
(i) download the highest available resolution for each URL and discard sources below the target resolution;
(ii) normalize all videos to $720$p, $24$\,fps, and a fixed $16{:}9$ center crop, removing letterboxing and black borders so that subsequent camera estimation operates on a clean field of view;
(iii) run scene detection to obtain single-shot segments;
(iv) split segments into fixed-length \(T\) videos via uniform temporal slicing.

We then filter videos with a single diagnostic based on motion.
Specifically, we run RAFT~\cite{teed2020raft} to estimate optical-flow magnitude; videos with too little motion are removed, while videos with stable, coherent motion are kept to emphasize informative, parallax-rich trajectories rather than static views.
Only videos satisfying the motion constraint enter the final dataset.
This yields a large-scale, in-the-wild drone corpus explicitly tailored to stress-test geometry-aware, trajectory-following video generation.

\begin{table*}[t]
\centering
\caption{
\textbf{Benchmark camera-controlled video generation.} \emph{Captain Safari} ranks first in 3D consistency and trajectory following with competitive video quality. Compared to the ablated variant without memory, \emph{Captain Safari} substantially improves 3D consistency and trajectory following, with only a slight trade-off in video quality. (Recon. = reconstruction rate. CosSim = cosine similarity.)
}
\vspace{-0.5em}
\label{tab:main_results}
\small
\renewcommand{\tabcolsep}{8.05pt}
\begin{tabular}{lcccccccc}
\toprule
\textbf{Model} & \multicolumn{2}{c}{\textbf{Video Quality}} & \multicolumn{2}{c}{\textbf{3D consistency}} & \multicolumn{3}{c}{\textbf{Trajectory Following}} \\
& FVD $\downarrow$ & LPIPS $\downarrow$ & MEt3R $\downarrow$ & Recon. $\uparrow$ & AUC@30 $\uparrow$ & AUC@15 $\uparrow$ & CosSim $\uparrow$ \\
\midrule
Geometry Forcing~\cite{wu2025geometry} &2662.75 &0.667 &0.4834 &0.877 &0.168 &0.056 &0.429 \\
Real-CamI2V~\cite{li2025realcam} &1585.61 &0.513 &\underline{0.3703} &\underline{0.923} &0.174 &0.051 &0.296 \\
Wan2.2-5B-Control-Camera~\cite{wan2025} &1387.75 &0.545 &0.3932 &0.767 &0.181 &0.054 &0.420 \\
Captain Safari w/o \textit{Mem.} &\textbf{998.47} &\textbf{0.504} &0.3720 &0.912 &\underline{0.193} &\textbf{0.068} &\underline{0.508} \\
Captain Safari &\underline{1023.46} &\underline{0.512} &\textbf{0.3690} &\textbf{0.968} &\textbf{0.200} &\textbf{0.068} &\textbf{0.563} \\
\midrule

\end{tabular}\vspace{-1em}
\end{table*}

\begin{table*}[t]
\centering
\caption{\textbf{Human preference.} Users overwhelmingly prefer \emph{Captain Safari} across all criteria, capturing \textbf{67\%} of total votes. The memory-removed variant ranks a distant second, while baselines competitors receive single-digit preference.}
\vspace{-0.5em}
\small
\label{tab:human}
\renewcommand{\tabcolsep}{15.3pt}
\begin{tabular}{lccc|c}
\toprule
\textbf{Model} & \textbf{Video Quality} & \textbf{3D consistency} & \textbf{Trajectory Following} & \textbf{Average} \\
\midrule
Geometry Forcing~\cite{wu2025geometry}   & 0.20\% & 0.00\% & 0.20\% & 0.13\% \\
Real-CamI2V~\cite{li2025realcam}         & 4.20\% & 6.40\% & 4.40\% & 5.00\% \\
Wan2.2-5B-Control-Camera~\cite{wan2025}  & 3.20\% & 3.80\% & 6.40\% & 4.47\% \\
Captain Safari w/o \textit{Mem.}         & 25.00\% & 24.20\% & 20.00\% & 23.07\% \\
Captain Safari                           & \textbf{67.40\%} & \textbf{65.60\%} & \textbf{69.00\%} & \textbf{67.33\%} \\
\midrule

\end{tabular}\vspace{-1em}
\end{table*}

\subsection{Camera Trajectory Reconstruction}

For each curated video, we estimate camera intrinsics and extrinsics at $4$\,fps using hloc~\cite{sarlin2019coarse,sarlin2020superglue}.
Specifically, we reconstruct an SfM model using COLMAP's incremental mapper with a Simple Radial camera model; we then export camera parameters as initial trajectories.

To obtain deployment-ready data, we apply a three-stage verification-and-fix pipeline to every reconstructed trajectory.
First, \emph{database check} consumes SfM statistics (inlier counts and ratios) to flag potentially unreliable transitions.
Next, \emph{geometric check} revisits suspicious pairs using stored keypoints and matches, recomputes essential matrices, and thresholds symmetric epipolar errors.
Last, \emph{kinematics check} analyzes the pose sequence for translation spikes, rotation jumps, forward-direction flips, and higher-order smoothness violations, using robust MAD-based scores to detect implausible motion.

The per-transition decisions are fused into a binary bad-index, which drives a strict policy. If bad transitions are sparse and localized, we invoke a targeted fix: we linearly interpolate camera centers and apply SLERP to rotations with a capped interpolation angle, optionally extrapolating at video boundaries. The fixed segments are then re-validated by the same database/geometric/kinematics criteria. If post-fix validation succeeds, the trajectory is exported into the final dataset. If the bad-index is too dense, violations are too severe, or fixed trajectories still fail verification, the entire video is discarded.

The resulting \emph{OpenSafari} couples high-dynamic, in-the-wild FPV drone video with rigorously verified camera trajectories.
It departs from existing benchmarks by emphasizing aggressive 6-DoF motion, strong parallax, and complex outdoor layouts, while enforcing strict geometric and kinematic validation.
This makes \emph{OpenSafari} a challenging testbed for camera-controllable video generation.

\begin{figure*}[t]
\centering
\includegraphics[width=\linewidth]{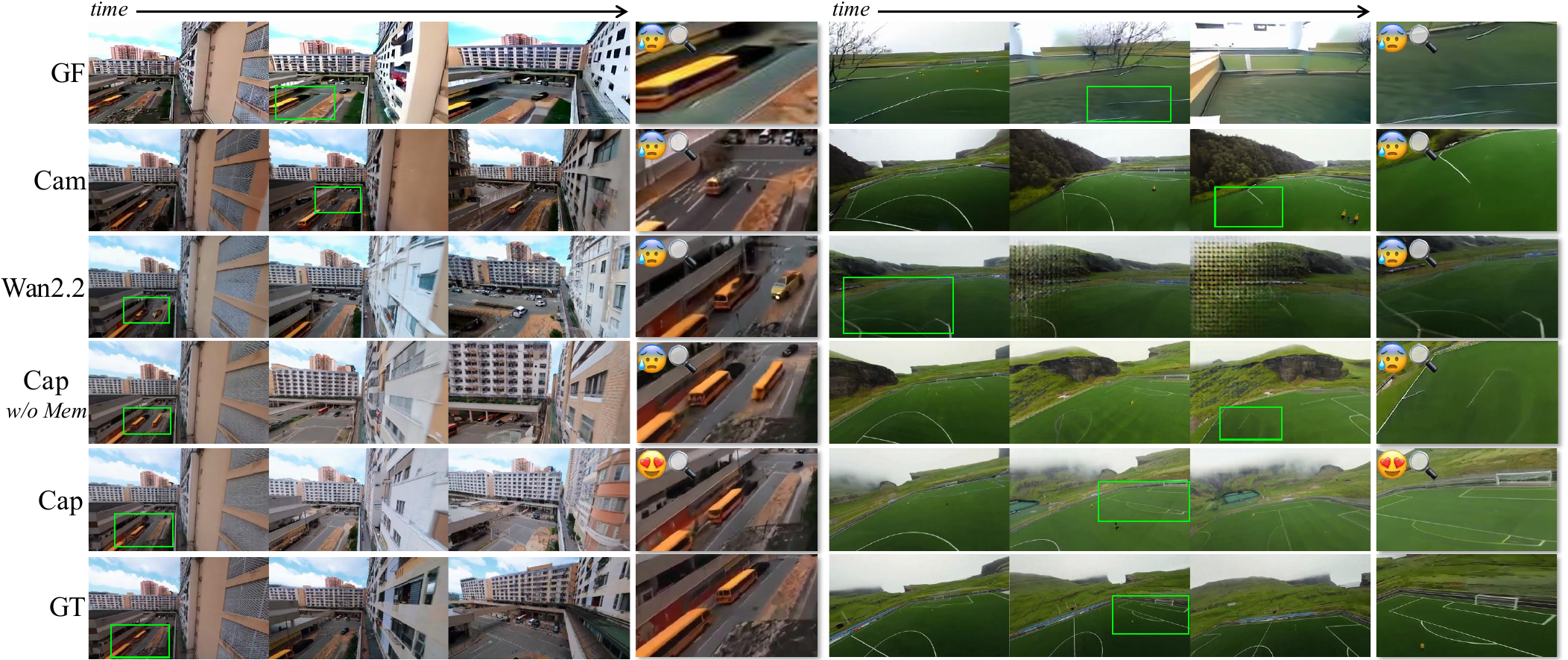}
\vspace{-1.5em}
\caption{\textbf{Qualitative comparisons.}
\textbf{Left}: Baselines—including the memory-removed variant—exhibit abrupt popping/vanishing of the school bus, and GF is low-quality. \emph{Captain Safari} alone renders the bus smoothly exiting the frame.
\textbf{Right}: Baselines distort or lose field marking, with Wan2.2 collapsing under large camera motion, affirming the challenge of 3D consistency under rapid trajectories. \emph{Captain Safari} preserves crisp markings and coherent layout while following the fast 6-DoF path (See Supp. Sec. \ref{sec:addi_res} for more qualitative results).\vspace{-1em}
}
\label{fig:qual}
\end{figure*}

\section{Experiments}

\subsection{Implementation Details}
\textbf{Training recipe.}
We adopt a two-stage recipe.
We first warm up the pose-conditioned memory retriever using pose-aligned memory tokens $m_t$.
We then jointly train the retriever and DiT end-to-end, updating the DiT via LoRA~\cite{hu2022lora}.
Memory cross-attention is initialized from the corresponding context cross-attention weights, and other new layers use standard initialization.

\noindent\textbf{Dataset.}
We extract overlapping clips with 1\,s stride, yielding $51{,}997$ training candidates.
A diversity-based trajectory filter removes clips with near-static motion, resulting in $11{,}481$ final training clips.
We additionally construct a non-overlapping test set of $787$ clips for evaluation.
For each clip, we generate a single descriptive caption using Qwen2.5-VL-7B~\cite{bai2025qwen2} and use it as the text condition.

\noindent\textbf{Configuration and notation.}
We generate $\mathcal{T}=5$\,s clips at 24\,fps from $T=15$\,s videos.
Camera poses and memory features are sampled at 4\,fps.
We use the terminal pose $p_{t_1}$ as the query as it represents the farthest viewpoint where drift typically accumulates, thereby strengthening geometric constraints across the entire trajectory.
The memory window is limited to $L=5$\,s.
We use Wan2.2-Fun-5B-Control-Camera~\cite{wan2025} as our base DiT with a hidden dimension $D = 3072$. 
Retriever and DiT are trained with $1$ and $5$ epochs, respectively.
Note that the retriever is also used during inference, and for simplicity and reproducibility, \(\mathcal{M}\) is constructed from the GT video.
We extract 3D-aware memory feature from StreamVGGT~\cite{zhuo2025streaming}.
We select four layers $\{4, 11, 17, 23\}$; at each layer, the feature contains $782$ tokens.
Concatenating across the four layers yields $M = 4 \times 782$ and $d_m = 1024$ memory tokens per frame.

\subsection{Benchmark}

\noindent\textbf{Metrics.}
We evaluate video generation along three complementary axes: video quality, 3D consistency, and trajectory following. For video quality, we report FVD~\citep{unterthiner2018towards} and LPIPS~\cite{zhang2018unreasonable}. For 3D consistency, we use MEt3R~\citep{asim2025met3r}, computed between GT and generated videos at matched time steps and a reconstruction rate that measures the percentage of frames successfully registered in the recovered 3D model~\citep{sarlin2019coarse,sarlin2020superglue}. For trajectory following, we report camera relocation accuracy (AUC~\citep{wang2025vggt}) and the cosine similarity between the flattened camera pose.
To evaluate generated videos, we apply the exact same SfM stack used for dataset curation, without a separate relocalization module. To handle monocular scale ambiguity, we canonicalize each video by fixing the first frame as the origin with identity rotation, ensuring AUC and CosSim evaluate relative extrinsics.

\noindent\textbf{Baselines.}
We compare against representative camera-controllable video generation models, including Geometry Forcing~\cite{wu2025geometry}, Real-CamI2V~\citep{cami2v,li2025realcam}, and Wan2.2-5B-Control-Camera~\cite{wan2025}, which cover geometry-constrained, reconstruction-driven, and large-scale diffusion-based approaches to trajectory-conditioned video synthesis.

\noindent\textbf{Human Study.}
We conduct a human study with 50 participants. Each participant is presented with 10 cases, where each case contains the GT video and five anonymized model-generated videos (three baselines, our model, and its ablated variant). For every case, participants are asked to select the best video under three criteria: Video Quality, 3D Consistency, and Trajectory Following. In total, the study collects $50 \times 10 \times 3=1,500$ human preference votes.

\subsection{Generation Quality}
\label{sec:gen_quality}
As shown in Table~\ref{tab:main_results}, our \emph{Captain Safari} attains a substantially lower FVD (1023.46 vs.\ 1387.75) and a slightly improved LPIPS score (0.512 vs.\ 0.513) compared to the SOTA baseline, demonstrating more stable temporal dynamics and sharper spatial details. Moreover, the human study in Table~\ref{tab:human} indicates that \textbf{67.40\%} of participants prefer our videos over all competing methods, highlighting the perceptual realism and overall fidelity of our generations.

Qualitative comparisons in Figure~\ref{fig:qual} further reveal that \emph{Captain Safari} produces visually compelling, realistic, and highly authentic scene dynamics. These findings are also consistent with the samples shown in Figure~\ref{fig:teaser}, where our method delivers vivid, coherent, and natural-looking drone videos that closely resemble real-world captures.

\begin{figure*}[t]
\centering
\includegraphics[width=0.9\linewidth]{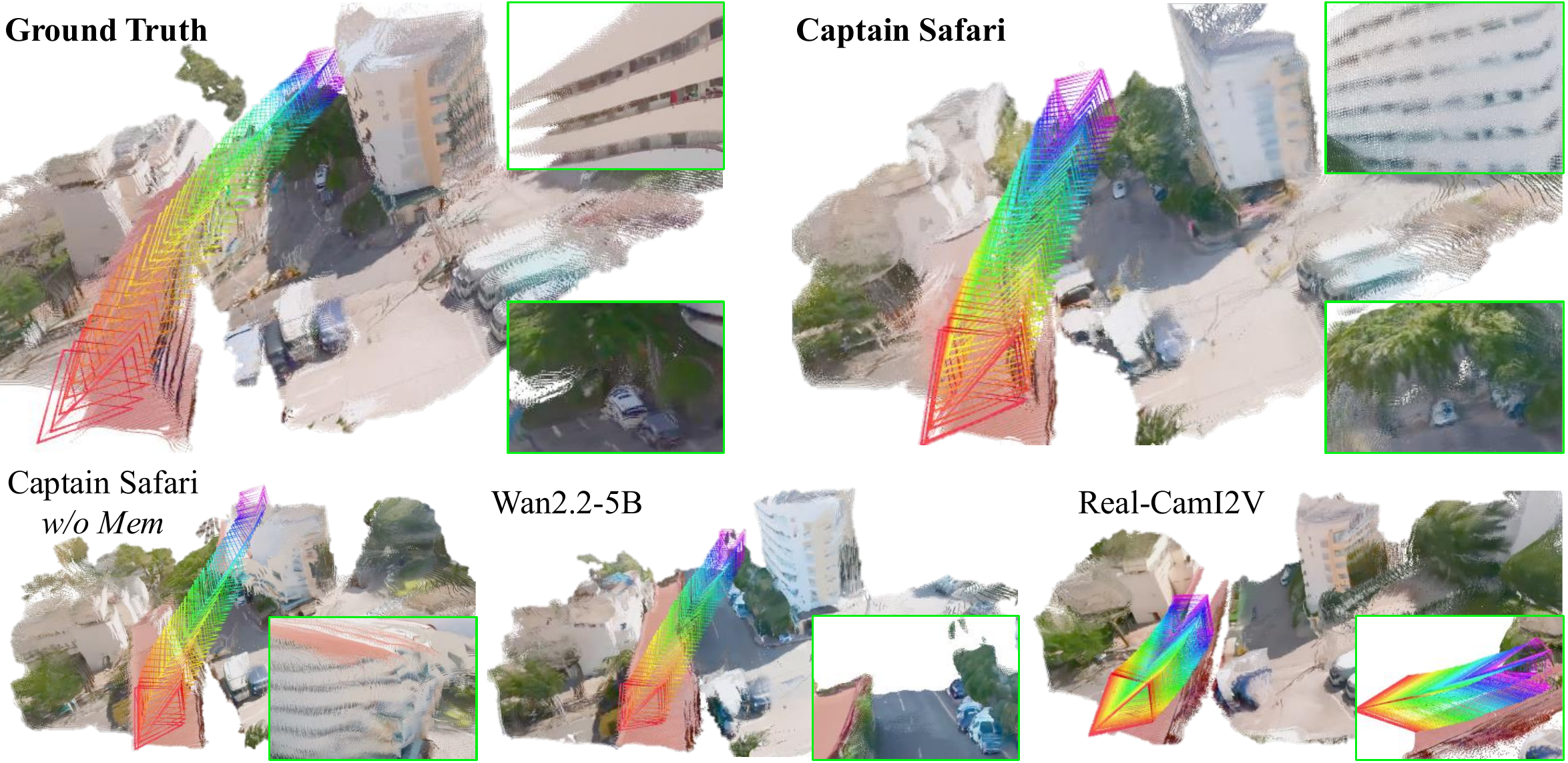}
\vspace{-0.3em}
\caption{\textbf{Scene reconstruction and camera trajectory.}
With pose-aligned memory, \emph{Captain Safari} reconstructs a well-structured building façade (the memory-removed variant blurs/warps it), demonstrating the benefit of memory. 
It also preserves fine details—parked cars and the tree on their roofs—that Wan2.2-5B fails to retain. 
Meanwhile, Real-CamI2V follows only a short path, whereas \emph{Captain Safari} covers the full trajectory with stable 3D structure, highlighting the challenge of maintaining 3D consistency under fast motion.\vspace{-1em}
}
\label{fig:reconstruction}
\end{figure*}

\subsection{3D Consistency}
\emph{Captain Safari} achieves state-of-the-art 3D consistency. As shown in Table~\ref{tab:main_results}, our method lowers MEt3R by 0.0013 (0.3690 vs.\ 0.3703, with a relative variance decrease of 10\%, Levene $p=0.0439$) and raises the reconstruction rate by 0.045 (0.968 vs.\ 0.923) copared to the strongest baseline. Consistently, the human study in Table~\ref{tab:human} shows that \textbf{65.60\%} of participants prefer \emph{Captain Safari} for 3D consistency, substantially surpassing all competing approaches.

Qualitative visualizations further confirm these quantitative gains. In Figure~\ref{fig:teaser}, structures such as the Greek-style columns remain geometrically stable across large viewpoint changes. 
In Figure~\ref{fig:qual}, our model produces (\textit{left}) a school bus that smoothly moves out of the frame, and (\textit{right}) preserves crisp, globally consistent field markings on the soccer pitch, whereas baselines exhibit distortions and disappearance. 
Figure~\ref{fig:reconstruction} and Figure~\ref{fig:teaser} further show that our reconstructions yield sharper façades and well-formed windows without collapsing geometry.
Together, these results validate that the implicit world memory and pose-conditioned retrieval of \emph{Captain Safari} effectively stabilize the underlying 3D world under aggressive camera motion.

\subsection{Trajectory Following}
\emph{Captain Safari} delivers the most accurate trajectory following among all competing models. As shown in Table~\ref{tab:main_results}, our method achieves the highest AUC@30 (0.200) and AUC@15 (0.068), along with the best cosine similarity (0.563), outperforming the strongest baseline by clear margins. The human study in Table~\ref{tab:human} further reinforces this observation, with \textbf{69.00\%} of participants identifying our model as the most faithful to the target camera path.

Figure~\ref{fig:reconstruction} provides a clear visualization of these improvements. \emph{Captain Safari}’s predicted trajectory closely aligns with the ground-truth path, while the ablated variant deviates and flies over the rooftop, and RealCam-I2V fails to follow the intended forward motion, advancing only slightly rather than committing to the prescribed trajectory. 
Furthermore, our method demonstrates stable generation under complex camera maneuvers in Figure~\ref{fig:teaser}. 
These results highlight the effectiveness of our memory-augmented, pose-conditioned design for precise trajectory adherence.

\subsection{Ablation Study}
Our results highlight the importance of the proposed pose-conditioned memory. As shown in Table~\ref{tab:main_results}, adding memory yields substantial improvements in both 3D consistency and trajectory following. These gains confirm that retrieving pose-aligned world tokens at the target frame provides the model with an explicit understanding of what the scene \emph{should} look like, enabling stable geometry and accurate motion alignment (see Supp. Table~\ref{tab:supp_ablation} for more ablations).

While removing memory (w/o \textit{Mem.}) slightly improves FVD and LPIPS, it severely degrades 3D consistency and trajectory following. This indicates a quality-consistency trade-off, where memory acts as a strong geometry prior that regularizes appearance optimization in favor of 3D stability. 
Qualitative comparisons in Figure~\ref{fig:qual} and Figure~\ref{fig:reconstruction} further illustrate these effects. With memory, the generated scenes preserve global structure, maintain consistent geometry across viewpoints. In contrast, the ablated variant often drifts and exhibits geometric inconsistencies. 
%

Furthermore, since we decouple memory retrieval from the DiT denoising loop to output a fixed-size \(w_t\), the computational overhead scales constantly (See Supp. Table \ref{tab:supp_complexity}).

\section{Conclusion}
We introduced \emph{Captain Safari}, a pose-conditioned world engine built on a world memory that enables long-range, 3D-consistent video generation under complex FPV trajectories. 
Together with \emph{OpenSafari}, our curated dataset of in-the-wild drone videos with verified camera poses, this establishes a rigorous benchmark for controllable video generation.
\emph{Captain Safari} markedly improves 3D consistency and trajectory accuracy over prior methods while maintaining strong visual fidelity. 
Future work will explore real-time world engines, as well as autoregressive generation where the global memory is continuously built from generated frames.
We hope \emph{Captain Safari} and \emph{OpenSafari} encourage further research in persistent world models and long-horizon controllable video generation.

\noindent \textbf{Acknowledgements.} This work is supported by the Office of Naval Research with N00014-23-1-2641.
Yu-Cheng Chou also acknowledges the computational resources provided by Cihang Xie and the advisement from Junfei Xiao.
{
    \small
    \bibliographystyle{ieeenat_fullname}
    \bibliography{main}
}

\clearpage
\setcounter{page}{1}
\maketitlesupplementary

\section{Additional Ablations on Memory Retrieval}
To further validate the design of our pose-conditioned memory retrieval, we compare our cross-attention-based retrieval with a simpler \emph{nearest-pose} baseline. In the nearest-pose variant, rather than aggregating pose-aligned world tokens via cross-attention, the model simply uses the memory feature $m_{\tau}$ corresponding to the past frame whose camera pose is closest to the query pose $p_t$. 

As shown in Table~\ref{tab:supp_ablation}, replacing our retrieval mechanism with nearest-pose memory degrades performance across all metrics compared to \emph{Captain Safari}. More importantly, the nearest-pose variant even underperforms the DiT baseline (w/o \emph{Mem.}). 
This is because a single nearest pose is often unaligned with the entire clip's viewpoint span under the aggressive 6-DoF motion of FPV drone flights, and this misalignment misguides the denoising process.

\begin{table}[h]
  \centering
  \caption{\textbf{Additional ablation on memory retriever.} A single nearest pose misguides the denoising process under aggressive FPV motion, resulting in worse performance than the DiT baseline (w/o \emph{Mem.}) and Captain Safari.}
  \vspace{-0.5em}
  \resizebox{\columnwidth}{!}{%
  \begin{tabular}{lcccccc}
  \toprule
    \multirow{2}{*}{\textbf{Ablation}} & \multicolumn{2}{c}{\textbf{Video Quality}} & \multicolumn{2}{c}{\textbf{3D Consistency}} & \multicolumn{2}{c}{\textbf{Trajectory Following}} \\
& FVD$\downarrow$ & LPIPS$\downarrow$ & MEt3R$\downarrow$ & Recon.$\uparrow$ & AUC@30$\uparrow$ & CosSim$\uparrow$  \\
    \midrule
    w/o $Mem.$ &\textbf{998.47} &\textbf{0.504} &\underline{0.3720} &\underline{0.912} &\underline{0.193} &\underline{0.508}\\
    Nearest-pose &1042.56 &0.517 &0.3810 &0.836 &0.185 &0.492\\
    \rowcolor{iblue!10} Captain Safari &\underline{1023.46} &\underline{0.512} &\textbf{0.3690} &\textbf{0.968} &\textbf{0.200} &\textbf{0.563} \\
    \bottomrule
  \end{tabular}}
  \vspace{-1em}
  \label{tab:supp_ablation}
\end{table}

\section{Complexity and Scalability Analysis}

To address potential concerns regarding the computational overhead of retrieving and conditioning on the world memory, we provide a detailed complexity analysis here.

A core advantage of \emph{Captain Safari} is its highly scalable design, which stems from \emph{decoupling} the memory retrieval process from the iterative DiT denoising loop. During inference, we first process the local memory $\mathcal{M}_{\text{local}}$ and query it in a \emph{single pass} to obtain the pose-aligned world tokens $w_t$. Because $w_t$ has a \emph{fixed size} (controlled by the number of learnable queries $M$), the subsequent denoising loop only cross-attends to this compact set of tokens.

Assuming generating a video sequence of length $\mathcal{T}$ requires $S$ denoising steps. Standard self-attention in a DiT costs $\mathcal{O}(S\mathcal{T}^2)$.
As shown in Table~\ref{tab:supp_complexity}, our formulation adds a one-time retrieval cost of $\mathcal{O}(|\mathcal{M}_{\text{local}}|)$, plus a cross-attention term $\mathcal{O}(S\mathcal{T}|w_t|)$ in the denoising loop. Since $|w_t|$ is fixed and very small compared to the full history, the cost of scaling the memory length $|\mathcal{M}_{\text{local}}|$ only impacts the one-time retrieval. Consequently, our computational scaling with respect to memory length is \emph{essentially constant} during the heavy iterative denoising phase.
This contrasts sharply with standard cross-attention approaches that attend to the full memory at every denoising step and scale linearly, as well as early-concatenation approaches that scale quadratically.

\begin{table}[h]
  \centering
  \caption{\textbf{Complexity comparison.} By decoupling memory retrieval and using fixed-size world tokens, our denoising overhead remains essentially constant with respect to $|\mathcal{M}_{\text{local}}|$.}
  \vspace{-0.5em}
  \resizebox{\columnwidth}{!}{%
  \begin{tabular}{lcccc}
  \toprule
    \textbf{Complexity} $\mathcal{O}(\cdot)$ & \textbf{DiT Base} & \textbf{Captain Safari} & \textbf{Full Cross-Attn} & \textbf{Concatenation} \\
    \midrule
    Retrieving cost & N/A & $|\mathcal{M}_{\text{lc}}|$ & N/A & N/A \\
    Denoising cost & $S\mathcal{T}^2$ & $S\mathcal{T}^2\!+\!S\mathcal{T}|w_t|$ & $S\mathcal{T}^2\!+\!S\mathcal{T}|\mathcal{M}_{\text{lc}}|$ & $S(\mathcal{T}\!+\!|\mathcal{M}_{\text{lc}}|)^2$\\
    \midrule
    \rowcolor{iblue!10} Scaling w.r.t $\mathcal{M}_{\text{lc}}$ & N/A & $\approx$ \textbf{Constant} & Linear & Quadratic \\
    \bottomrule
  \end{tabular}}
  \vspace{-1em}
  \label{tab:supp_complexity}
\end{table}

\section{Dynamic Scene Generation and Additional Qualitative Results}\label{sec:addi_res}
One concern regarding world engines is their ability to model the real world, which is inherently dynamic rather than purely static. While our implicit memory and trajectory formulation provide a strong geometric prior for stable backgrounds, our generator is built on top of powerful diffusion priors. Consequently, \emph{Captain Safari} goes beyond rigid 3D rendering and remains highly capable of synthesizing dynamic content. 

As demonstrated in our supplementary materials\footnote{\href{https://johnson111788.github.io/captain-safari-supp/}{\texttt{johnson111788.github.io/captain-safari-supp/}}}, the model can generate plausible moving elements such as moving vehicles and natural phenomena (e.g., ocean waves) with stable 3D environments.

For extensive additional qualitative results, including high-resolution video comparisons, and demonstrations of dynamic capabilities under aggressive 6-DoF motion, strong parallax, and complex outdoor layouts, please visit our project page\footnote{\href{https://johnson111788.github.io/open-safari/}{\texttt{johnson111788.github.io/open-safari/}}} and the supplementary website\footnote{
\href{https://johnson111788.github.io/captain-safari-supp/}{\texttt{johnson111788.github.io/captain-safari-supp/}}}.

\section{Data Usage and Ethics}
To comply with platform policies (e.g., AirVuz and YouTube) and respect copyright ownership, we do not redistribute the raw video files. Instead, \emph{OpenSafari} is released as a comprehensive, open-source data curation pipeline. We provide the full suite of downloading, preprocessing, semantic filtering, and camera reconstruction scripts\footnote{\href{https://github.com/johnson111788/Captain-Safari/}{\texttt{github.com/johnson111788/Captain-Safari/}}}. Researchers can use this pipeline to fetch the original URLs and automatically reproduce our carefully verified, geometry-annotated dataset for non-commercial, academic research purposes.

\end{document}